# Towards a comprehensive taxonomy of online abusive language informed by machine leaning

Samaneh Hosseini Moghaddam[a,*], Kelly Lyons, Cheryl Regehr[a], Vivek Goel[b], Kaitlyn Regehr[c]

[a]*University of Toronto, Toronto, Canada*
[b]*University of Waterloo, Waterloo, Canada*
[c]*University College London, London, UK*

**Abstract**

The proliferation of abusive language in online communications has posed significant risks to the health and wellbeing of individuals and communities. The growing concern regarding online abuse and its consequences necessitates methods for identifying and mitigating harmful content and facilitating continuous monitoring, moderation, and early intervention. This paper presents a taxonomy for distinguishing key characteristics of abusive language within online text. Our approach uses a systematic method for taxonomy development, integrating classification systems of 18 existing multi-label datasets to capture key characteristics relevant to online abusive language classification. The resulting taxonomy is hierarchical and faceted, comprising 5 categories and 17 dimensions. It classifies various facets of online abuse, including context, target, intensity, directness, and theme of abuse. This shared understanding can lead to more cohesive efforts, facilitate knowledge exchange, and accelerate progress in the field of online abuse detection and mitigation among researchers, policy makers, online platform owners, and other stakeholders.

*Corresponding author
  Email address:* `samaneh.moghaddam@utoronto.ca` (Samaneh Hosseini Moghaddam)

# 1. Introduction

In recent years, the proliferation of abusive language in online communications has posed significant risks to individuals and communities, leading to social and psychological harm (El Asam and Samara, 2016; Wachs et al., 2020; Weber et al., 2020a). Research in this area has focused on cyberbullying among school age children and youth (Cassidy et al., 2013; Mishna et al., 2016; Patchin and Hinduja, 2022) and post-secondary students (Davani et al., 2022; Mishna et al., 2019; Whittaker and Kowalski, 2015), followed by research addressing abuse, harassment and threats of individuals in a wide variety of occupational groups including health care professionals (D'Souza et al., 2017; Larkin, 2021; Regehr et al., 2023; Symons et al., 2021; Ward et al., 2022), teachers (Dolev-Cohen and Levkovich, 2021; Kopecký and Szotkowski, 2017; Kyriacou and Zuin, 2016), social service workers (Alfandari et al., 2023; Burns et al., 2024; Kagan et al., 2018) and journalists. While online harassment and threats alone can have consequences for those who experience it (Oksanen et al., 2022; Park and Choi, 2019; Privitera and Campbell, 2009; Regehr et al., 2023; Symons et al., 2021), there is increasing evidence that violence does not remain confined to the digital realm but can manifest in the real world, contribute to in-person harassment and physical violence (Gover et al., 2020; Nguyen, 2023; Regehr et al., 2024; Weber et al., 2020b). The growing concern necessitates methods for identifying and mitigating harmful content and facilitating continuous monitoring, moderation, and early intervention (Rawat et al., 2024).

Over the past decade, a wide range of abusive language detection approaches have been developed (Alrashidi et al., 2022; Ejaz et al., 2024; Gandhi et al., 2024; Haidar et al., 2016; Iwendi et al., 2023; Mansur et al., 2023; Poletto et al., 2021; Salawu et al., 2017; Yin and Zubiaga, 2021), accompanied by labeled datasets needed for training machine learning detection models (Albanyan et al., 2023; Albanyan and Blanco, 2022; Almohaimeed et al., 2023; Kennedy et al., 2022; Toraman et al., 2022; Vidgen et al., 2021a). Abusive language manifests in various forms, from explicit threatening hate speech to more subtle forms of verbal aggression. It is a complex and multifaceted concept that has proven difficult to recognize, both by humans and machines (Poletto et al., 2021). To capture the nuances of these diverse expressions of harm, a multi-dimensional classification approach is essential, as it allows for a better understanding of the spectrum of online abuse (Bianchi et al., 2022). As a result, many detection models leverage multi-label datasets wherein samples are annotated to indicate not only whether the text is abusive or not, but also to consider various facets such as the target, type, and theme of abuse.

However, there is a lack of consensus on a standardized taxonomy for classifying abusive language, particularly regarding the categories and dimensions used. Existing datasets use diverse and non-standard classification systems in their labeling schemes, which consequently leads to detection models providing inconsistent outputs. In some cases, labels may be applied in contradictory ways. For example, Toraman et al. (2022) labels abusive language as "hate speech" if it incites violence or is threatening while other forms of abuse such as humiliation, discrimination, and insults are labeled as "offensive language ". In other datasets (Kirk et al., 2021; Sachdeva et al., 2022; Vidgen et al., 2021b), all these forms of abuse are classified as "hate speech". Given legal definitions of hate speech in countries such as Canada, where it falls under the Criminal Code, this particular term is fraught with complexity and controversy (Department of Justice, 1985). Additionally, in some datasets, the specific subject of abuse is called a "target" (Kennedy et al., 2022; Kirk et al., 2021) but in other datasets, the subject of the abuse is labeled as "topic" (Bourgeade et al., 2023), "theme" (Salminen et al., 2018), or "domain" (Toraman et al., 2022).

The incompatibilities in the classification structure and terminology poses challenges in re-purposing data, comparing outcomes, integrating datasets, and ensuring study generalizability. Furthermore, it complicates communication and knowledge sharing among experts. Poletto et al. (2021) highlight that hate speech in text often encompasses multiple interconnected phenomena, yet current research frequently addresses only one or limited isolated aspects. This narrow focus limits understanding of the broader context of abusive language. Therefore, developing a shared, data-driven taxonomy is essential, as it can clarify the relationships and distinctions among these complex concepts. A unified framework would not only enhance collaboration among researchers but also facilitate more comprehensive studies that address the multifaceted nature of hate speech and related phenomena.

Several taxonomies have been proposed and used to categorize abusive texts, either explicitly as part of a research study's contribution or implicitly as a labeling scheme for a labeled dataset (Lewandowska-Tomaszczyk, 2022; Salminen et al., 2018; Vidgen et al., 2021a). However, they cover only a subset of the facets of abusive language and no widely accepted comprehensive taxonomy exists that encompasses the various facets. A comprehensive taxonomy is essential to enable consistent and comparative analyses across studies, disciplines, and contexts. It not only supports re-usability of artifacts such as datasets and computational models, but also facilitates cross-disciplinary integration, fostering the cumulative development of knowledge. Moreover, to our knowledge, none of the existing classification systems have been developed following a systematic methodology or evaluated rigorously, leading to redundancy and imbalanced granularity. This lack of a comprehensive, evaluated taxonomy often makes existing frameworks less reliable and challenging to apply consistently across studies that focus on varying dimensions of abuse.

In this paper, we propose a comprehensive taxonomy of abusive language as a standardized classification framework. This taxonomy is developed systematically by reconciling key characteristics identified by previous researchers in the field of abusive language detection. Our methodology is based on the Nickerson-Varshney-Muntermann (NVM) method (Nickerson et al., 2013), which is a well-known and widely accepted taxonomy development method (Szopinski et al., 2019). We adopt and extend the hierarchical version of NVM (Nickerson et al., 2024) to capture the complexity and conceptual breakdown of abusive language across multiple dimensions and respective characteristics. The systematic development ensures the suitability of the structure of the taxonomy which is assessed and demonstrated in Section 6.

An novel aspect of our approach is that we use existing labeled abusive language datasets from past research to identify relevant categories, dimensions, and characteristics. By utilizing labels from diverse datasets across various abusive language detection systems, our taxonomy captures a broad range of abusive language dimensions. We leverage the definitions and relationships among the labels within existing datasets as a rich conceptual resource, grounded in the empirical observations made by the researchers who curate these datasets. Although the use of a wide range of classification systems makes the proposed taxonomy highly general, its faceted structure [1] ensures flexibility, allowing for future extensions without requiring significant changes to the taxonomy's overall framework.

---

[1] A faceted taxonomy, in contrast to ordinary strict hierarchical taxonomies, is a classification system that organizes information through multiple independent dimensions, known as facets. It offers high flexibility by allowing the incorporation of new terms without altering the overall structure.(Ferreira et al., 2017).

The proposed taxonomy offers a framework for mapping labels from existing datasets to the nodes of the taxonomy, establishing a common terminology and clear associations between the labels of datasets and the classes of classifiers. This mapping facilitates the comparison and merging of datasets, while also promoting effective communication and knowledge sharing.

We evaluate the suitability of the structure of our taxonomy using metrics proposed by Bedford (2013): generality, appropriateness, and orthogonality. Generality measures the granularity of labels. A taxonomy should capture necessary detail without being overly complex. Appropriateness assesses completeness and soundness to determine how thoroughly the taxonomy accounts for relevant concepts. Orthogonality assesses the degree of independence and non-overlap among the labels.

## 2. Background

Clarifying and unifying a single definition of abusive language is a critical step in developing an effective taxonomy for its detection. This section presents a review of existing taxonomies and related work and discusses foundational concepts and definitions of abusive language upon which our taxonomy is developed.

### 2.1. Definition of abusive language

In online communications, researchers widely agree that there is no consensus on what constitutes abusive language and other forms of harmful language (Founta et al., 2018; Seemann et al., 2023; Waseem et al., 2017). The categorization of abusive content relies on subjective and theoretical criteria, which cannot be reduced to legal definitions or platform guidelines, as these are often minimalistic, vague, or influenced by commercial interests. (Vidgen et al., 2019). There are differences and overlaps among definitions in various papers that consider forms of harmful language (Waseem et al., 2017).

Caselli et al. (2020) define abusive language as "hurtful language that a speaker uses to insult or offend another individual or a group of individuals based on their personal qualities, appearance, social status, opinions, statements, or actions", and claims that their definition is more comprehensive than previous definitions. Similarly, Nobata et al. (2016) use the term "abusive language" to refer to hurtful language including hate speech, derogatory language, and profanity. In their study the term "abusive language" is used as an umbrella term to encompass different forms of harmful texts that sometimes are referred to by cyberbullying, hate speech, toxic, offensive language, and online aggression.

In alignment with these broader perspectives, we define abusive language as any form of hurtful language that harms, insults, demeans, or intimidates individuals or groups. This can include explicit or implicit references to personal characteristics such as race, gender, sexuality, religion, or other traits, with the potential of causing emotional distress or encouraging physical violence.

In our definition of abuse, we consider only that which is directed towards individuals or groups. In addition to individual- and group-directed abuse, Vidgen et al. (2019) identified abuse that is directed against a concept such as belief systems, countries, or ideologies. Concept-directed abuse allows for the expression of critical opinions and dissenting perspectives on various topics, contributing to constructive debate and dialogue without necessarily harming individuals or groups. Since we are focused on abusive language directly targeting humans, we do not include concept-directed abuse in our definition.

## 2.2. Classification Systems for Abusive Language

Although numerous studies have suggested classification systems for abusive language, either explicitly by introducing a new taxonomy or implicitly by designing a labeling scheme, as far as we know, none of these systems has been claimed to be developed and evaluated using a systematic method. Consequently, none of these classification systems have been demonstrated to be comprehensive to fully cover the diverse classes required by different dataset labeling schemes.

Waseem et al. (2017) proposed a classification system that captures similarities and differences among aspects of abusive language. They distinguish between individuals and groups as targets and between implicit and explicit hate. Zampieri et al. (2019) employed a hierarchical annotation schema that distinguishes between offensive and non-offensive tweets, targeted and untargeted if they are offensive, and the target type such as individual or group for those identified as targeted.

Vidgen et al. (2021a) presented a three-level taxonomy for classifying abusive language. The first level differentiates between abusive or non-abusive language. Subsequently, abusive language is further categorized into three distinct classes: identity-directed, affiliation-directed, and person-directed, based on the target of the abuse. The first two categories, identity- and affiliation-directed, focus on the severity and directness of abusive language, with sub-classifications for each category: derogation and animosity, threatening language, glorification, and dehumanization. In terms of directness, animosity is considered implicit, while the other classes are deemed explicit. In the third category, person-directed abuse is delineated between abuse directed towards someone actively engaged in the conversation and abuse directed towards individuals not participating but clearly identified within the discourse. Non-abusive language is subdivided into non-hateful slurs, counter-speech, and neutral expressions. Moreover, counter-speech samples are categorized into three additional categories based on the target of the abuse they aimed to counter including against identity-, affiliation-, and person-directed abuse.

As far as we know, Salminen et al. (2018) presented the only taxonomy for abusive language that can be considered a faceted taxonomy. Their taxonomy for hateful online comments includes language of hate (accusations, promoting violence, humiliation, or swearing) and target of abuse (financial power, political issues, racism, religion, specific nation, specific person, media, armed forces, or behaviour). In most definitions of abusive language or even hate speech, hate targeting media or behaviour is not considered within the scope of abusive language or even hateful comments because these do not target an individual or a group of people (Kennedy et al., 2022; Kirk et al., 2021; Mathew et al., 2021; Röttger et al., 2020). Another challenge with this taxonomy is that the sub-groups within the target of hate are not homogeneous or of the same type. For example, specific person category specifies if the text is targeting an individual or a group of people while categories such as race, religion, and nationality are all considered different varieties of hate. Moreover, the taxonomy is incomplete as it does not cover several prevalent themes of abuse, such as gender and sexual orientation and language such as threatening.

The taxonomy we propose in this paper is a general-purpose taxonomy designed to categorize general abusive language similar to the taxonomies reviewed thus far. Some studies have focused on developing taxonomies for specific types of abuse or specific classes of abusive language. Among them, gender-based abuse has received the most attention (Fersini et al., 2018; Zeinert et al., 2021; Kurrek et al., 2020). Others have focused on racist attitudes (ElSherief et al., 2021), and religion (Ramponi et al., 2022).

It is noteworthy that some papers provide taxonomies to be used as the labeling scheme of datasets in languages other than English (Fortuna et al., 2019; Liebeskind et al., 2023; Lu et al., 2023; Vidgen and Derczynski, 2020). All datasets that were used in the process of developing our proposed taxonomy are in English. However, the proposed taxonomy could potentially be used as a standard framework in other languages.

## 3. Datasets

This section outlines the process of preparing a list of available labeled abusive language detection datasets for use in developing the taxonomy based on their labeling schemes. Performing a systematic review to curate a precise list of published datasets in the area of abusive language detection is significantly costly and out of the scope of this study; thus, we used two indices of abusive language datasets provided by two different research groups. One is a catalog and comparison of 59 English-language datasets that Vidgen and Derczynski (2020) have curated and maintain named "Hate speech catalog" [2]. The other is a collection of 46 English-language abusive language datasets maintained by Piot et al. (2024)[3] named "Meta hate". Both collections include datasets published between 2016 and 2023. We merged these two catalogs and filtered them based on the type and source of samples to include datasets relevant to the domain of this research in the following ways.

Our focus is on textual general-purpose English language datasets collected from human communications in online websites and social media platforms, while excluding multi-media (Suryawanshi et al., 2020), specific datasets (Gautam et al., 2020; Grimminger and Klinger, 2021; Kurrek et al., 2020; Vidgen et al., 2020), datasets in languages other than English, datasets involving bots (Cercas Curry et al., 2021), and non-online sources (Saker et al., 2023; Sarker et al., 2023). Another out of scope dataset is one by Pamungkas et al. (2020) where the focus is on the contribution of swear words to abusive text, as the annotation level in such datasets is on individual swear words rather than entire texts. After completing all refinements, the final list of datasets was arranged in reverse chronological order based on their publication year. Table 1 presents some details about the final 18 datasets used in our study.

As indicated in Table 1, there are inconsistencies in the number and types of labels used in the different datasets. For example, the CounterHate dataset uses 4 binary labels, while the MHS dataset incorporates a much more complex structure with 11 numerical and 53 binary labels. This variation highlights the lack of standardization in defining and categorizing labels across datasets. To address these inconsistencies, we performed the following preprocessing steps on the labeling schemes of the datasets:

- Dealing with Noisy Data: In cases where annotators are allowed to either select themes of abuse from a predefined list or suggest new ones, we can encounter instances of redundancy and misspelling in the labels. To address these issues, we standardize themes by identifying and merging redundant or misspelled labels.

- Recombining Annotations with Majority Voting: In datasets where detailed annotations from multiple annotators are provided instead of final labels, we applied majority voting to determine the labels. This approach helped consolidate conflicting annotations into a single, co-hesive label, improving the reliability and consistency of the labeling scheme. Majority voting, in which each annotator's label is treated as a vote and the label with the most votes is selected as the final decision, is a widely adopted and straightforward technique for resolving annotation disagreements, as it effectively captures the most commonly agreed-upon label while minimizing the impact of outlier judgments (Davani et al., 2022).

- Breaking down Compound Labels: When datasets contain compound labels representing multiple characteristics of abusive language, we decompose these into individual labels to enable more precise categorization and analysis. This approach ensures that each characteristic is distinctly identified and analyzed.

- Flattening the Labels: To ensure consistency across datasets, we simplified hierarchical or multi-level labels into a single level. For instance, in the CAD dataset, labels indicating the intensity of hate, such as "Animosity," "Dehumanization," and "Derogation," are repeated under different categories such as "Affiliation" and "Identity." By flattening these labels, we unify similar characteristics into a single label and consistent category.

---

[2] https://hatespeechdata.com/
[3] https://irlab.org/metahate.html

| Order | Dataset Name | Citation | Year | Source of Data | Size | Number and Types of Labels |
|---|---|---|---|---|---|---|
| 1 | CounterHate | (Albanyan et al., 2023) | 2023 | Twitter | 2K | 4 binary |
| 2 | HateComment | (Gupta et al., 2023) | 2023 | YouTube, BitChute | 2K | 1 binary, 1 categorical |
| 3 | HSData | (Mody et al., 2023) | 2023 | Mixed online content, social media, and synthetic | 726K | 1 binary |
| 4 | Toraman | (Toraman et al., 2022) | 2022 | Twitter | 100K | 1 categorical |
| 5 | Albanyan | (Albanyan and Blanco, 2022) | 2022 | Twitter | 2K | 4 binary |
| 6 | MHS | (Sachdeva et al., 2022) | 2022 | YouTube, Reddit, Twitter | 39K | 11 numerical, 53 binary |
| 7 | GabHC | (Kennedy et al., 2022) | 2022 | Gab | 27K | 14 binary |
| 8 | CAD | (Vidgen et al., 2021a) | 2021 | Reddit | 25K | 2 categorical |
| 9 | Semeval | (Pavlopoulos et al., 2021) | 2021 | Civil Comments | 10K | 2 binary (annotators' votes) |
| 10 | Dynabench | (Vidgen et al., 2021b) | 2021 | Synthetic | 41K | 1 binary, 2 categorical |
| 11 | Hatemoji | (Kirk et al., 2021) | 2021 | Synthetic | 5K | 1 binary, 2 categorical |
| 12 | HateXplain | (Mathew et al., 2021) | 2021 | Twitter, Gab | 20K | 1 categorical, 1 list of categorical (annotators' votes) |
| 13 | CONAN | (Fanton et al., 2021) | 2021 | Synthetic | 5K | 2 categorical |
| 14 | Ethos | (Mollas et al., 2020) | 2020 | YouTube, Reddit | <1K | 8 numerical |
| 15 | Toxicity | (Pavlopoulos et al., 2020) | 2020 | Wikipedia | 10K | 2 categorical (annotators' votes) |
| 16 | AbuseEval | (Caselli et al., 2020) | 2020 | Twitter | 14K | 1 categorical |
| 17 | SWAD | (Pamungkas et al., 2020) | 2020 | Twitter | 1K | 1 binary |
| 18 | Hatecheck | (Röttger et al., 2020) | 2020 | Synthetic | 3K | 1 binary, 2 categorical |

Table 1: The first 18 datasets from merging the two collections, "Hate Speech Catalog" and "Meta Hate," sorted by publication year.

  Additionally, we standardized all labels by transforming them into binary formats. Specifically, categorical labels were converted into binary labels by representing the presence or absence of each category as 1 or 0, respectively. For numerical labels, we determined a midpoint threshold based on the range of values in each dataset. Values above the midpoint were encoded as 1, and those below or equal to the midpoint were encoded as 0.

- Adding Common Characteristics: In some datasets, there are certain characteristics that are universally applicable to all samples, thus they are not explicitly provided as labels. For instance, in the Albanyan dataset where all samples represent reactions to hate speech, there is no specific label indicating whether a text is an original post or a reaction to a post. To ensure that these implicit characteristics are recognized and consistently represented, we explicitly add them as labels to the samples in the datasets.

- Ignoring Irrelevant Labels: To maintain the focus and relevance of our

taxonomy, we excluded labels that do not align with our definition of abusive language. For example, in the HateComment dataset, concept-directed labels such as all instances of "Organization" (e.g. WHO, McDonalds, and US Government) were disregarded. This step was essential to ensure that only pertinent characteristics of abusive language were included, thereby enhancing the specificity and applicability of our taxonomy.

## 4. Methodology

A taxonomy is a structured classification system that organizes entities into classes based on shared key characteristics. We adopt and extend the hierarchical taxonomy development method proposed by Nickerson et al. (2024). It is a goal-based methodology that begins by defining the domain and purpose of the taxonomy and then employs an iterative process to identify and refine relevant taxonomy elements, ensuring each is aligned with the pre-defined goal. The identification of taxonomy elements can be performed either conceptually (primarily derived from existing domain knowledge), or empirically (by investigating a subset of the entities being studied).

As the goal of our study is to develop a taxonomy that can be used to identify and analyze abusive language, the domain of our taxonomy is abusive language, and the purpose is to classify online texts by distinguishing abusive language and its various facets. Publicly available abusive language datasets used as the basis of computational abusive language models are a rich source of domain knowledge grounded in empirical observations. Each dataset includes a labeling scheme designed to address specific aspects of online abuse detection, offering valuable insights into the relevant key characteristics and distinctions of abusive language.

To capture the complexity of abusive language and its various facets using the datasets as our source of knowledge, we extended both the definition and process of the NVM method proposed by Nickerson et al. (2013) as described in the following two subsections.

### 4.1. Taxonomy Definition

Nickerson et al. identified three fundamental concepts that constitute a taxonomy: categories, dimensions, and characteristics, allowing for a hierarchical structure (Nickerson et al., 2024). Specifically, they define a hierarchical taxonomy $T_h$ as a collection of categories $Cat_l$ where a category is a set of other categories $Cat_{li}$ and/or dimensions $D_{li}$ and a dimension $D_{li}$ is a set of characteristics $C_{kij}$ as illustrated in Equation 1 (Nickerson et al., 2024). This definition classifies entities within a hierarchical structure. Dimensions represent attributes of the entities and can be applied at various levels within the hierarchy of categories. Each dimension consists of characteristics that are mutually exclusive and collectively exhaustive. Moreover, each entity has one and only one characteristic for each dimension.

$$\text{Th} = \big\{\text{Cat}_\ell, \ell = 1, \ldots, L \,\big|\, \text{Cat}_\ell = \big\{\text{Cat}_{\ell i} \vee D_{\ell i}, i = 1, \ldots, I \,\big|\, \\ D_{\ell i} = \big\{\text{Cl}_{ij}, j = 1, \ldots, J_i, J_i \geq 2\big\}\big\}\big\} \qquad (1)$$

There is a noteworthy difference between dimensions, categories, and characteristics. Entities are classified into categories. Dimensions define a set of characteristics that are used to describe and differentiate entities within categories of the taxonomy. Unlike categories and characteristics, which describe specific attributes or properties of entities within the taxonomy, dimensions identify a specific hierarchy level at which their corresponding characteristics are relevant. In other words, dimensions act as organizational layers that group related characteristics, while categories and characteristics provide detailed information about the entities. For example, in a taxonomy of abusive language, a dimension might represent a broad aspect such as "Target of abuse" and its characteristics might specify attributes such as abuse targeting "Race" or "Gender", which are relevant to entities within the category of abusive language but are not applicable to non-abusive language texts. In what follows, we refer to all components of a taxonomy including categories, dimensions, and characteristics as *elements* and the categories and characteristics as *classes*.

We extend the definition of a taxonomy used in the NVM method by permitting a hierarchical structure within both dimensions and characteristics while preserving the conceptual relationships between categories and dimensions, as well as between dimensions and characteristics. The extended definition is presented in Equation 2. Each dimension may be either a meta-dimension or a basic dimension. A meta-dimension comprises a list of dimensions, while a basic dimension consists of a set of characteristics. Characteristics under each basic dimension are organized in a hierarchical structure across one or more levels. At each level, the characteristics are mutually exclusive and collectively exhaustive. Furthermore, each entity is assigned one and only one characteristic at each level for each dimension.

$$\text{Th} = \big\{\text{Cat}_\ell, \ell = 1, \ldots, L \mid \text{Cat}_\ell = \big\{\text{Cat}_{\ell i} \vee D_{\ell i}, i = 1, \ldots, I \mid D_{\ell i} = \big\{D_{\ell ij}, j = 1, \ldots, J_i, J_i \geq 2\big\} \vee \big\{\text{Cl}_{\ell ij}, j = 1, \ldots, J_i, J_i \geq 2 \mid \text{Cl}_{\ell ij} = \big\{\text{Cl}_{\ell ij}\big\} \vee \big\{\text{Cl}_{\ell ijk}, k = 1, \ldots, K_{ij}, K_{ij} \geq 2\big\}\big\} \quad (2)$$

This extension enhances our ability to model the complexity of abusive and non-abusive texts and their various facets. For instance, the theme of abuse is a dimension of abusive texts. It comprises several dimensions, including gender, race, and religion. Since an abusive text may be relevant to multiple themes, such as "East Asian women", we cannot simply categorize them as characteristics of a single dimension. Instead, we must classify them in separate dimensions, all of which are conceptually relevant. The use of hierarchical dimensions allows us to represent these relationships.

Furthermore, the granularity of the characteristics within a dimension may vary across different contexts. For example, one context may require distinguishing between abuse directed at an individual versus that directed at a community. In another context, it may be crucial to differentiate between targeting an individual who is party to the communication and one who is not. The hierarchical arrangement of the characteristics allows modeling various level of granularity.

*4.2. Taxonomy Development Methodology*

In the NVM method, a hierarchical taxonomy is developed in two separate iterative phases: (1) the hierarchical structure of the taxonomy, which consists of all categories and sub-categories is iteratively determined, (2) the dimensions and characteristics are identified in a separate iteration. While it is feasible to create a taxonomy for abusive language using this process, we believe that developing categories alongside other elements—dimensions and characteristics— enables consideration of the interactions among all elements at once, making the process more straightforward.

Moreover, the NVM method suggests that in both phases, the taxonomy elements are identified either conceptually or empirically. We use the labeling schemes from available datasets as encoded domain knowledge and apply the labeled data as empirical support for identifying the elements of the taxonomy. These labeling schemes are rooted in empirical observations by other researchers in the field. As a result, in our taxonomy, elements are identified through a combination of conceptual reasoning and empirical evidence.

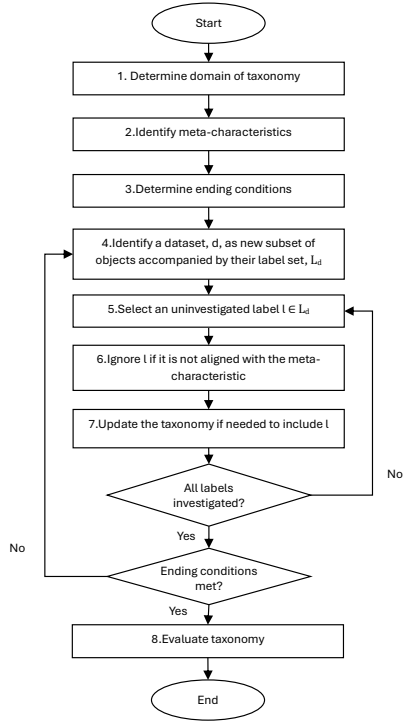

Figure 1: The taxonomy development process which is an extension of the NVM method (Nickerson et al., 2024)

Our taxonomy development method is illustrated in Figure 1. To support inclusion decisions for characteristics, the first two steps determine of the domain and the meta-characteristic for the taxonomy. The goal of our study is to develop a taxonomy that can be used to identify and analyze abusive language, thus the domain of our taxonomy is abusive language text. The meta-characteristic acts as an overarching trait that forms the foundation for selecting characteristics within the taxonomy. Each characteristic should logically stem from the meta-characteristic, ensuring coherence and relevance in the classification process. The meta-characteristic identified from the goal and domain of the taxonomy is, "Aspects of online texts that contribute to the study of abusive language texts".

The next step is to determine the ending conditions. The NVM method makes use of objective and subjective ending conditions. Objective ending conditions ensure that taxonomy development is complete, for example, by examining whether any new dimension or characteristic is added to the taxonomy in the last iteration. Subjective ending conditions track the quality of the taxonomy and examine if the taxonomy is concise, robust, comprehensive, extendable, and explanatory. We adopted all of the objective and subjective ending conditions from the NVM method and added the following ending condition to ensure that the process will not stop before all datasets published in a given year have been examined: No new element is added to the taxonomy after investigating all datasets in the last published year.

After deciding on ending conditions, the taxonomy is developed through two nested loops: the outer loop iterates over a collection of existing datasets, while the inner loop iterates over labels from the dataset under investigation. Each label is either mapped to an existing element of the taxonomy or is used to update the taxonomy to incorporate the concept of the label into the taxonomy. Although taxonomy development involves qualitative judgments, we use the following specific process, rationale and criteria in the development of the proposed taxonomy:

- For each label, we first review the paper that introduces the dataset and describes its labels. Additionally, we search for any supplementary artifacts, such as annotation guides or GitHub repositories, for further consultation.

- To test our perception of the label definition and its boundaries—especially for labels with limited descriptive information—we fetch and review a few labeled examples from the dataset. This allows us to assess how the label is applied in practice and refine our understanding of its scope based on concrete instances, rather than solely relying on textual definitions.

- The label is investigated if it is relevant to the meta-characteristic and only relevant labels are considered for inclusion in the taxonomy. The relevancy of a label is identified using the information collected through step 1 and 2.

- Beginning at the root of the existing taxonomy hierarchy, we systematically traverse each node, ensuring that the labels are mapped to the most specific applicable category or characteristic. Considering the subjective ending condition, where a label does not fit within any existing element or the label is a well-defined subset of the mapped element, we introduce new elements to represent the concept. This may involve defining new categories, dimensions, or characteristics consistent with the NVM framework.

Following this method allows us to gradually identify and refine categories and key characteristics of different dimensions of abusive language, and incrementally develop and refine the elements of the taxonomy considering new labels. We also used the following general principles that guided the development of the taxonomy in step 4. These principles are meant to ensure that the taxonomy remained clear, consistent, and adaptable to evolving trends in abusive language.

- Recognizability by General Public: As abusive language is usually generated by people in the general population, it is essential that the terms and classes used in the development of the taxonomy can be understandable and distinguishable by the public. It enhances its usability, particularly when dealing with concepts that might be confusing or interchangeable for non-experts.

- Granularity Consistency: The taxonomy should maintain a consistent level of detail across all classes, avoiding any disproportionate representation or imbalance in the specificity of categories.

- Long-Term Relevancy: Certain targets of abuse are dynamic and time dependent, being strongly dependent on global events and recent news. For example, specific nationalities or professional affiliations are relevant targets of abuse at times when certain events are happening (such as wars or a pandemic). In developing our taxonomy, we deliberately avoid creating categories of specific nationalities or professional affiliations from labels in the studied datasets.

## 5. Taxonomy of Abusive Language

After following the methodology using the labeling schemes of available datasets, a taxonomy for online abusive language was developed. The iterative development process was repeated until all ending conditions were met, at which point the process was concluded. There was a total of eighteen iterations resulting in eighteen datasets and 119 labels being examined. The datasets are listed in Table 1. The final taxonomy is organized into 5 categories, 17 dimensions and 100 characteristics as shown in Figure 2. Categories and characteristics are represented by rectangles, while dimensions are depicted as ovals. Additionally, in this figure, all characteristics of each dimension of the "Theme of Abuse" are listed within a single shared rectangle for the sake of space management.

Our taxonomy is a hierarchical faceted taxonomy. The primary advantage of a faceted taxonomy is its flexibility(Kwasnik, 1999). For instance, abusive language can be classified by target (e.g. individual, group), intensity (e.g. threatening, derogatory), directness (e.g. explicit, implicit), and theme of abuse (e.g. religion, gender). As a result, a single abusive text sample can be classified as threatening, targeting an individual, explicit, and based on religion. Furthermore, the adaptability of faceted taxonomies makes its evolution straight-forward. As new insights emerge, researchers can integrate additional labels into the taxonomy without disrupting the existing framework. Thus, faceted taxonomies enhance collaboration among researchers and across disciplines.

Each dimension of the taxonomy, represents a facet of online text or abusive language. Under each dimension, its corresponding characteristics are organized. In some cases, the characteristics are arranged in a hierarchical structure such that the characteristics in each level are mutually exclusive and collectively exhaustive. Furthermore, the union of the ending nodes across all levels constitutes a collection of all possible characteristics associated with that dimension.

The first level of the taxonomy consists of one dimension,"Context" and two categories,"Non-abusive", and "Abusive" with further subdivisions under each. "Context" is a hierarchical concept that applies to all kinds of online texts. Following the taxonomy development process, the "Context" dimension was identified in the available datasets as having two levels of granularity. In the first level, the classification determines whether the online text is an original post or a reaction to another post. If the text is categorized as a reaction, it is further classified into three subclasses based on the nature of the original post it responds to: "To Abusive", "To Non-abusive", and "To Counter Hate".

"Non-Hateful Slurs" as defined in the CAD dataset taxonomy by Vidgen et al. (2021a) refer to pejorative terms that do not always indicate identity-directed abuse. They can be explicitly insulting or implicitly express animosity but are sometimes used to counter prejudice or have been reclaimed by a targeted group. Since there can be hateful and non-hateful slurs, this category should be used to classify non-hateful uses of slurs, while hateful uses should be classified as identity-directed abuse.

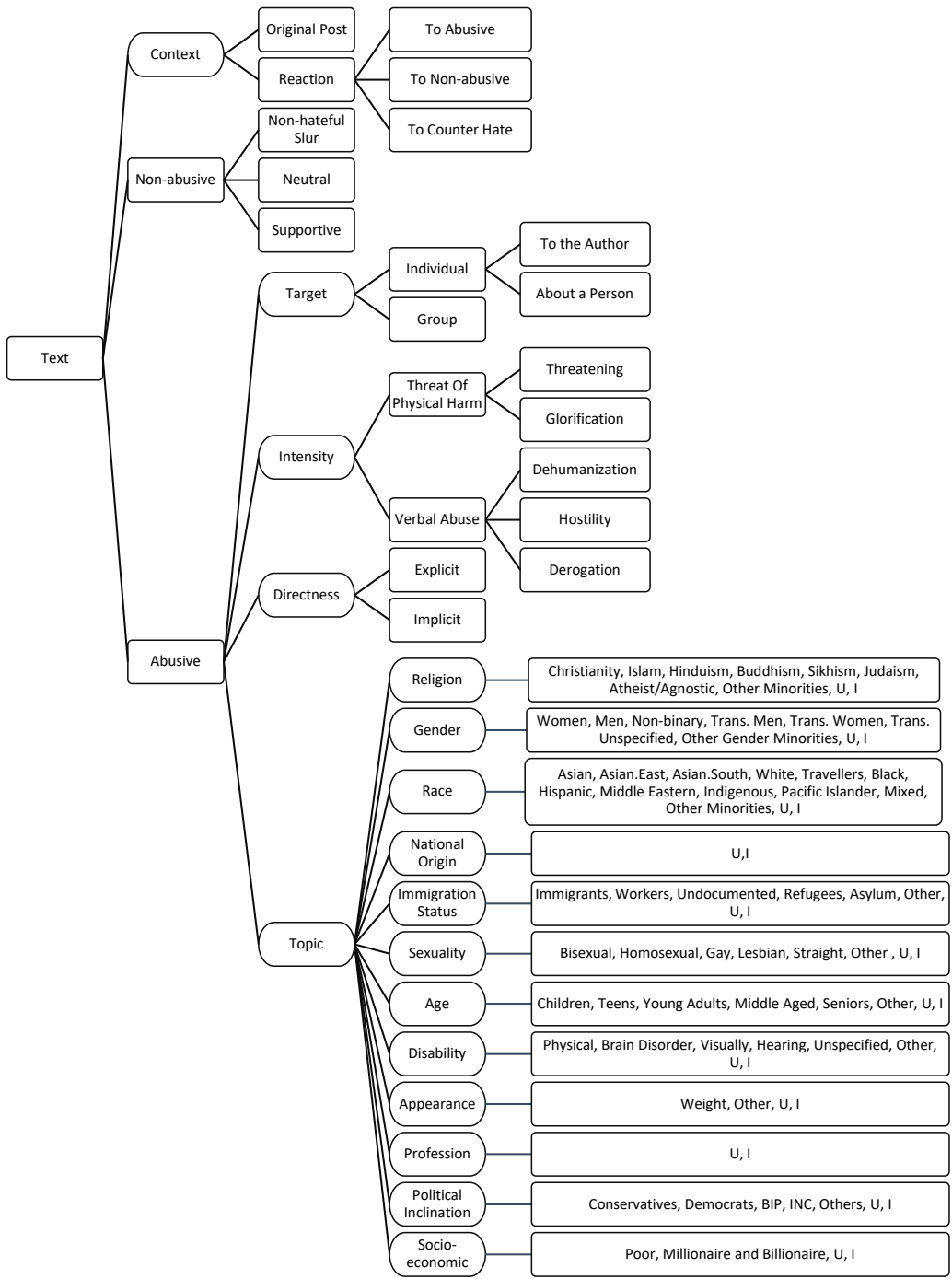

Figure 2: Proposed Taxonomy for Abusive Language (U = Unspecified, I = Irrelevant)

"Supportive" language refers to positive comments in support of identifiable communities of people. In some datasets, "Counter Speech" (Vidgen et al., 2021a) or "Counter Hate" (Albanyan et al., 2023; Albanyan and Blanco, 2022) labels are introduced that distinguish a form of communication that challenges, condemns, or calls out abusive language. As a result, text that contains hateful content directed at the author of the original hate speech may be labeled as "Counter Speech" or "Counter Hate". According to our definition of abuse, any type of harmful language, regardless of intent, is classified as abusive. For example, in any controversial discussion about two sides of a war, all hateful speech, whether it is original content or a negative reaction to abusive content should be categorized as abusive in our taxonomy. We acknowledge that this definition does not align with Vidgen et al. (2021a)'s taxonomy which categorizes "Counter Speech" as non-abusive even though it might contain hateful content. In general, the category "Supportive" language in our taxonomy should be used to classify only non-hateful text that counters hate. Finally, "Neutral" content encompasses all non-abusive text that is neither non-hateful slurs nor supportive text.

The Abusive category is further divided into four primary dimensions that identify different facets of abusive language: (1) "Target", (2) "Intensity", (3) "Directness", and (4) "Theme" of abuse. Each will be discussed more in the following subsections.

## 5.1. Target

The "Target" dimension specifies whether the abusive language is directed at an individual or a group. For individuals, the person must be clearly identified, either by their actual name, username, or position. It further categorizes whether the abuse is directed toward the author or another person. "To the author" abuse is aimed directly at the content creator in a "Reaction" context; otherwise, the individual-directed abuse is considered to be "About a person": either it refers to someone other than the content creator in a reaction context or it mentions a specific person in an original context. This definition is close to what is provided in the CAD dataset, which is the only dataset that includes these two labels.

If abusive language is directed at a collective entity, affecting multiple individuals who share a common characteristic, the "Target" of abuse is a "Group". Common characteristics might include identifiable groups such as race and gender, professions, ideologies, affiliations, or other traits. The specific characteristic is not important; what matters is that the abuse is aimed at a group of people.

*5.2. Intensity*

The "Intensity" dimension of abusive language plays a role in determining the severity of abusive content. The first level of the dimension includes two characteristics: "Threat of physical violence" and "Verbal Abuse".

The "Threat of physical violence" characteristic involves language that implies physical harm, either through explicit threats of violence or the glorification of violent acts. This type of abuse is considered particularly severe as it presents a clear danger to the victim's physical well-being. Consequently, this characteristic is broken into two sub-characteristics: (1) "Threatening" and (2) "Glorification". "Threatening" text includes explicit threats of physical harm, intimidation towards violence, and even mentions of acts as severe as genocide. Language in this subcategory directly communicates an intention to cause physical harm to the victim. For example, explicit statements such as "I will kill you" or "You deserve to be exterminated" fall under this category. "Glorification" involves incitement to violence and the praising or celebrating of violent acts or harm. Text classified under this sub-characteristic endorses or encourages violent behaviour, thus indirectly posing a threat by promoting a culture of violence. For instance, statements that express admiration for historical figures associated with violence or advocate for aggressive actions are categorized as glorification of violence in our taxonomy.

On the other hand, "Verbal abuse" encompasses non-physical forms of abuse that are aimed at causing emotional harm. This characteristic includes behaviours that are intended to belittle, demean, or intimidate the target. "Verbal abuse" can have long-lasting psychological effects on the victim and is often used as a tool for exerting power and control over others. This characteristic includes three sub-characteristics: (1) "Dehumanization", (2) "Derogation", and (3) "Animosity".

"Dehumanization" includes language that denies the humanity of individuals or groups, reducing or comparing them to sub-human status such as objects or animals.

"Derogation" encompasses language that seeks to undermine, pigeonhole, or propagate stereotypes about individuals or groups. It involves reducing individuals to simplistic labels or caricatures, often based on perceived characteristics or affiliations. This type of abuse perpetuates harmful narratives and reinforces systemic biases, aiming to marginalize or humiliate those targeted.

"Animosity" includes language that accuses, condemns, or assigns blame to individuals or groups with a hostile or antagonistic tone. It involves casting aspersions, making unfounded allegations, or engaging in character assassination to tarnish reputations or incite conflict. This type of abuse fuels resentment, fosters division, and undermines trust within communities or relationships.

Differentiating these sub-characteristics helps in better identifying and addressing the varying levels of harm presented by different types of abusive content. Identifying the intensity of abuse could assist in determining the level of harm and consequently developing more effective strategies for detecting and combating harmful speech.

5.3. Directness

The "Directness" dimension distinguishes whether the abusive language is "Explicit" or "Implicit", highlighting the overt or subtle nature of the abuse. It indicates how abusive language is articulated (Vidgen et al., 2019). While explicit abuse is easier to identify and address, implicit abuse requires more nuanced and tailored approaches to detect (ElSherief et al., 2021; Ocampo et al., 2023).

"Implicit" abusive language is subtle and indirect, often relying on context, insinuation, or underlying tones to convey harmful or derogatory messages. It may involve innuendo, coded language, or passive-aggressive remarks that require interpretation to fully understand the abusive intent. This form of abuse can be particularly insidious because it is less obvious and can be easily dismissed or overlooked. A comment like "Someone of your background must find this task challenging" implies inferiority without directly stating it.

"Explicit" abusive language is overt and direct, leaving no room for ambiguity about its harmful intent. It involves clear, straightforward statements or actions meant to demean, insult, or harm another person or group. This form of abuse is easily recognizable and often includes offensive slurs, threats, or openly hostile remarks. A statement like "You are worthless and don't deserve to be here" clearly conveys abuse without any need for interpretation.

*5.4. Theme*

The "Theme" of abuse addresses social aspects targeted by abusive language, including religion, gender, race, national origin, immigration status, sexuality, age, disability, appearance, profession, political inclination, and socio-economic status. It specifies the characteristics of the theme of the abusive language, regardless of whether the target is an individual or a group.

As online abusive language can pertain to more than one theme of abuse, the recognized categories are arranged in a hierarchy of dimensions. Representing them as characteristics of the theme of abuse would violate the main principle that characteristics of each dimension should be mutually exclusive. As Figure 2 shows, each subdimension of the dimension "Theme" is associated with a collection of characteristics. Each collection of characteristics includes "U" and "I," representing unspecified and irrelevant, respectively. The inclusion of irrelevant ensures adherence to the principle that the characteristics within each dimension are collectively exhaustive. Since the lists of countries for national origin and professions or affiliations as themes of abuse are dynamic and subject to change over time, it is impractical and unnecessary to constrain the taxonomy by enumerating their specific characteristics.

Identifiable groups including religion, gender, race, national origin, immigration, sexual orientation, age, and disability as well as appearance are generally applicable sub-dimensions of theme of abuse for various applications of abusive language detection. In addition to general themes of abuse that are applicable to general-purpose abusive language detection applications, our taxonomy includes three application-specific themes of abuse: (1) "Political inclination" (Grimminger and Klinger, 2021), (2) "Profession/affiliation" (Charitidis et al., 2020; Gray et al., 2017), and (3) "Socio-economic" (Gupta et al., 2023).

*5.5. Mapping Between Labels and Taxonomy Classes*

During the process of taxonomy development, we documented how the labels of the analyzed datasets were mapped to classes of the taxonomy. Table 2 illustrates which datasets correspond to each class of the taxonomy. Due to space limitations, the characteristics within the abuse theme are consolidated. The "Iteration No." row in the table indicates the iteration in which each class of the taxonomy was initially added. By referencing the iteration number, it is possible to trace back to the dataset whose labeling scheme prompted the introduction of that class.

In this table, dark gray cells highlight explicit labels directly associated with each taxonomy class, and light gray cells represent labels inferred from the meta-data of the dataset such as source of data or the is-a relations in the hierarchy of the taxonomy. For example, in the CONAN dataset, there is no label indicating whether samples are abusive since it does not include any non-abusive samples. Consequently, the label "Abusive" is not explicitly provided, and the cell representing the intersection of CONAN and the "Abusive" class is shaded light gray to reflect this. As the context of samples is identical across all samples in each dataset, their labeling schemes do not include context. We identified the context of each dataset by investigating their descriptions and their source of data discussed in the corresponding paper. The HSData dataset is the only exception wherein there is a mixture of original posts and reactions. Therefore, these instances are also marked with light gray cells. Additionally, categories or characteristics in the taxonomy that have labels mapped to at least one of their sub-categories or sub-characteristics are shaded light gray, as we can infer the presence of positive samples based on sub-categories or sub-characteristics.

## 6. Evaluation

In the field of online abusive language classification taxonomies are typically developed without formal assessments of their effectiveness or suitability. To address this gap and demonstrate the appropriateness of our taxonomy, we conducted a detailed evaluation process. Kaplan et al. (2022) raised three perspectives for the assessment of a taxonomy's structural suitability: generality, appropriateness, and orthogonality. Moreover, they defined seven metrics for evaluating them, as detailed in Figure 3.

In the following sections, we define each of the metrics and show how they were used to evaluate the taxonomy. To define these metrics formally, we introduce the following notations: let $T$ represent our taxonomy and $C$ be the set of classes (categories, dimensions, and characteristics) in the taxonomy. We denote each class as $c \in C$. $D$ denotes the set of datasets under study, and for each dataset $d \in D$, $L_d$ represents the set of labels in $d$. Considering each dataset $d$, the set of all mappings between labels in $d$, $L_d$, and relevant classes in $C$ is denoted with $m_{L_d}^C \subseteq C \times L_d$. A label, $l \in L_d$. $D$, is associated with a class, $c \in C$, if $(c, l) \in m_{L_d}^C$.

| | Original post | Reaction | To abusive | To non-abusive | To Counter Hate | Abusive | Individual | To the author | About a person | Group | Threat of physical violence | Threatening | Glorification | Verbal abuse | Dehumanization | Hostility | Derogation | Explicit | Implicit | Religion | Race | National Origin | Immigration Status | Gender | Sexuality | Age | Disability | Appearance | Profession/Affiliation | Political inclination | Socio-economic | Non-hateful Slur | Neutral | Supportive |
|---|---|---|---|---|---|---|---|---|---|---|---|---|---|---|---|---|---|---|---|---|---|---|---|---|---|---|---|---|---|---|---|---|---|---|
| Iteration No. | 4 | 1 | 5 | 5 | 1 | 1 | 2 | 8 | 8 | 2 | 4 | 6 | 8 | 4 | 6 | 6 | 6 | 7 | 7 | 2 | 2 | 2 | 2 | 2 | 2 | 2 | 2 | 2 | 2 | 2 | 2 | 8 | 6 | 1 |
| CounterHate | | | | | | | | | | | | | | | | | | | | | | | | | | | | | | | | | | |
| HateComment | | | | | | | | | | | | | | | | | | | | | | | | | | | | | | | | | | |
| HSData | | | | | | | | | | | | | | | | | | | | | | | | | | | | | | | | | | |
| Toraman | | | | | | | | | | | | | | | | | | | | | | | | | | | | | | | | | | |
| Albanyan | | | | | | | | | | | | | | | | | | | | | | | | | | | | | | | | | | |
| MHS | | | | | | | | | | | | | | | | | | | | | | | | | | | | | | | | | | |
| GabHC | | | | | | | | | | | | | | | | | | | | | | | | | | | | | | | | | | |
| CAD | | | | | | | | | | | | | | | | | | | | | | | | | | | | | | | | | | |
| Semeval | | | | | | | | | | | | | | | | | | | | | | | | | | | | | | | | | | |
| Dynabench | | | | | | | | | | | | | | | | | | | | | | | | | | | | | | | | | | |
| Hatemoji | | | | | | | | | | | | | | | | | | | | | | | | | | | | | | | | | | |
| HateXplain | | | | | | | | | | | | | | | | | | | | | | | | | | | | | | | | | | |
| Ethos | | | | | | | | | | | | | | | | | | | | | | | | | | | | | | | | | | |
| CONAN | | | | | | | | | | | | | | | | | | | | | | | | | | | | | | | | | | |
| Toxicity | | | | | | | | | | | | | | | | | | | | | | | | | | | | | | | | | | |
| AbuseEval | | | | | | | | | | | | | | | | | | | | | | | | | | | | | | | | | | |
| SWAD | | | | | | | | | | | | | | | | | | | | | | | | | | | | | | | | | | |
| Hatecheck | | | | | | | | | | | | | | | | | | | | | | | | | | | | | | | | | | |

Table 2: Mapping Between Labels of Datasets and Taxonomy elements.

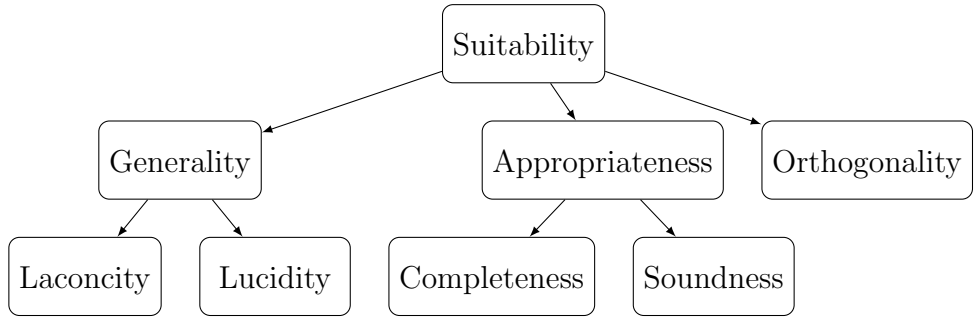

Figure 3: The break down of evaluation metrics

*6.1. Generality*

The metrics, laconicity and lucidity are used to measure how well the taxonomy strikes a balance between being too general or too specific in its classification.

A label, l, is considered laconic with respect to a taxonomy, $T$, if it is associated with at most one class in the taxonomy:

$$Laconic(T, L_d, l) = \begin{cases} 1 & if \left|\{c \mid (c, l) \in m_{L_d}^C\}\right| \leq 1 \\ 0 & \text{otherwise} \end{cases} \quad (3)$$

$Laconic(T, L_d, l)$ yields 1 if a label is laconic and 0 otherwise. The overall laconicity of the taxonomy is evaluated as an average of the laconic function of all labels in the datasets, defined as follows:

$$Laconicity(T, D) = \frac{\bm{\Sigma}_{d \in D} \bm{\Sigma}_{l \in L_d} Laconic(T, L_d, l)}{\bm{\Sigma}_{d \in D} |L_d|} \quad (4)$$

The resulting value lies between 0 and 1, where a higher laconicity value indicates that classes of the taxonomy are well-defined and not redundant across multiple labels, suggesting that the taxonomy is neither too fine-grained nor redundant.

Lucidity evaluates whether the classes in the taxonomy are lucid, meaning whether each class is associated with at most one label in each dataset. The lucid function, $Lucid(T, L_d, c)$, yields 1 if the class, $c$, is lucid and 0 otherwise. It is defined as follows:

$$Lucid(T, L_d, c) = \begin{cases} 1 & if \left|\{l \mid (c, l) \in m_{L_d}^C\}\right| \leq 1 \\ 0 & \text{otherwise} \end{cases} \quad (5)$$

The overall lucidity of the taxonomy is calculated as the average between the minimum lucidity score for each class c across all datasets. It is defined as follows:

$$Licidity(T, D) = \frac{\bm{\Sigma}_{c \in C} \left(\mathbf{min}_{d \in D} Lucid(T, L_d, c)\right)}{|C|} \quad (6)$$

The resulting value lies between 0 and 1, where a higher value indicates a higher proportion of lucid classes meaning that classes in the taxonomy are clearly defined by unique relevant terms, suggesting the taxonomy is appropriately specific and not overly broad or ambiguous.

A suitable generality in our taxonomy will be indicated by sufficiently high values for both laconicity and lucidity, demonstrating that the taxonomy is neither too specific nor too generalized. Lower values will help us identify potential issues in the taxonomy's structure and refine it for better clarity and granularity.

For the studied datasets, the computed laconicity and lucidity values are 1 and 0.85, respectively. The laconicity score is particularly strong, reflecting the effective generalization of our taxonomy. While the lucidity score is slightly lower, it remains sufficiently high to ensure that the taxonomy effectively distinguishes between the majority of classes. To better understand the factors contributing to the lower lucidity score, we examined all taxonomy classes where multiple labels from a dataset were mapped to the same category. The main reasons for the score reduction are summarized below:

- Broad Categories: As discussed in Section 4, we avoided adding subcategories for "National Origin" and "Affiliation" to ensure long-term relevance. Consequently, all related subcategories in the datasets were mapped to their respective superclasses.

- Overlapping Labels: Some datasets, such as MHS dataset, contain conceptually overlapping labels (e.g., Sentiment, Respect, Insult, Attack/Defend, Humiliate, Status, Dehumanization, Violence, and Genocide). These overlaps resulted in multiple labels being mapped to the same class, reducing lucidity. To preserve class independence, we excluded overlapping labels from our taxonomy.

- Interchangeable Terms: Certain labels, particularly in abusive language contexts, are not easily distinguishable by ordinary people and may be used interchangeably. For example, 'Black' and 'African' in Hatemoji dataset and 'cognitive disability' and 'neurological disability' in MHS dataset were consolidated in our taxonomy.

- Uneven classes: For the sake of granularity consistency, we avoid including uneven classes such as 'Nazis' in Dynabench dataset which label abusive language referencing Nazi values. Including such classes would introduce inconsistencies, as similar stereotypes and common forms of hate speech are not explicitly classified in our taxonomy.

In summary, the lower lucidity score in our taxonomy can be attributed to the intentional qualifications applied during the taxonomy development process, as discussed in Section 4. These decisions, while necessary for maintaining consistency and long-term relevance, have led to trade-offs in lucidity.

## 6.2. Appropriateness

Completeness and soundness are metrics for assessing a taxonomy's appropriateness, determining whether it fully and correctly covers all labels of the datasets under study. These metrics measure whether the taxonomy strikes a balance between comprehensive coverage and the avoidance of unnecessary classes.

Completeness is measured by the fraction of complete labels over all objects under study. A label, $l$, is considered complete with respect to a taxonomy, $T$, if there is at least one class $c$ in the taxonomy such that is associated with $l$. The completeness function is defined as follows:

$$Complete(T, L_d, l) = \begin{cases} 1 & if \left|\{c \mid (c,l) \in m_{L_d}^C\}\right| \geq 1 \\ 0 & \text{otherwise} \end{cases} \quad (7)$$

This function yields 1 if a label is complete and 0 otherwise. The overall completeness of the taxonomy is evaluated as an average of the Complete function of all labels in the datasets as defined as follows:

$$Completeness(T, D) = \frac{\Sigma_{d \in D} \Sigma_{l \in L_d} Complete(T, L_d, l)}{\Sigma_{d \in D} |L_d|} \quad (8)$$

The resulting value lies between 0 and 1, where a higher completeness value indicates that labels are well covered by the classes of the taxonomy, suggesting that there are no classes that should be added to the taxonomy. Soundness is measured by the fraction of sound classes in the taxonomy. A class $c$ is considered sound if there is at least one label associated with it. It is defined as:

$$Sound(T, L_d, c) = \begin{cases} 1 & if \left|\{l \mid (c,l) \in m_{L_d}^C\}\right| \geq 1 \\ 0 & \text{otherwise} \end{cases} \quad (9)$$

The overall soundness of the taxonomy is calculated as the average between the maximum sound score for each class $c$ across all datasets, $D$. It is defined as:

$$Soundness(T, D) = \frac{\Sigma_{c \in C} \left(\mathbf{max}_{d \in D} Sound(T, L_d, c)\right)}{|C|} \quad (10)$$

The resulting value lies between 0 and 1, where a low soundness indicates that the taxonomy may include unnecessary classes that can be removed or that the datasets under study lack diversity. Overall, appropriateness is ensured when both completeness and soundness are sufficiently high, indicating that the taxonomy fully and correctly covers all relevant labels without including unnecessary classes. The computed values for the completeness and soundness scores for the taxonomy are 0.99 and 0.96, respectively. While these values are strong, we did investigate very few cases that caused a slight drop in these metrics.

The completeness score is slightly impacted due to the inability to map just two labels from the CounterHate dataset— "Agrees with the counterhate tweet" and "Additional Counterhate when it agrees with the counterhate tweet"—to any class in our taxonomy. This is because some counterhate text contains abusive language that attack back the author of initial abusive language, making it difficult to definitively categorize the instances as non-abusive. Although this affects the completeness score, we decided not to address this issue by adding additional classes, as doing so could reduce the overall accuracy of the taxonomy.

Although our taxonomy development methodology is designed to incorporate new elements based on the observation of dataset labels, there are a few classes in the taxonomy that do not have associated labels in the datasets under study. These classes were incorporated to ensure the taxonomy adheres to the principle of collective exhaustiveness, as emphasized by NVM (Nickerson et al., 2013), which requires covering all relevant aspects of the concept, even if certain elements are not explicitly represented in the datasets. For instance, while the datasets include the label "Millionaire and Billionaire", which was categorized under the "Socio-Economic Factors" dimension, there is no complementary label such as "Poor" in the examined datasets. This contributes to the slight drop in soundness.

*6.3. Orthogonality*

Orthogonality requires that no two classes within a taxonomy overlap or share identical scope and boundaries. It helps ensure that the taxonomy maintains independent and separate classes. The recommended assessment method is to consider dependency between every two classes in the taxonomy using a self-referencing orthogonality matrix. After comparing every two classes of our taxonomy, the only potential overlap identified is between "Reaction" and "To the author". While every instance of "To the author" can also be classified under "Reaction" since it inherently involves the context of a post being an original or a repost, the two classes do not share identical scope or boundaries. "Reaction" refers to the nature of the post's origin—whether it is an original post or a reaction to a post—while "To the author," as a branch of target of an "Individual," distinguishes the target of abuse, specifying whether it is directed at the author of the post or someone external to the thread. Thus, although there is a dependency between the two, they represent distinct dimensions and do not violate the orthogonality principle, as their scopes remain independent and separate.

## 7. Practical Applications

A primary application of our taxonomy is to facilitate dataset integration. Effective abusive language detection systems require substantial amounts of labeled data to capture a broad spectrum of abusive language, from implicit derogatory stereotypes to explicit calls for violence. Creating such datasets, including data collection and annotation, is costly. Therefore, integrating and reusing existing datasets is a valuable solution, though often challenging due to differences in labelling schemes. Our taxonomy addresses this by providing a standardized framework that harmonizes diverse datasets. By mapping various datasets to the taxonomy, researchers and practitioners can create larger, more diverse training datasets and improve the generalizability of models trained on the datasets. Moreover, this standardization facilitates the comparison of models across studies and datasets, promoting more robust validation and benchmarking practices.

Our taxonomy offers a well-defined framework for labeling abusive language datasets. By providing a comprehensive set of categories and facets, the taxonomy assists researchers in designing annotation schema with greater consistency and detail. This consistency improves the usability of resulting datasets.

The hierarchical and faceted nature of our taxonomy supports a nuanced understanding of abusive language. This structure allows researchers to interpret model outputs with greater precision by breaking down abusive language into specific dimensions such as type, target, intensity, and context. By analyzing these dimensions, researchers can gain deeper insights into the underlying patterns. Additionally, it could be useful in facilitating more targeted interventions and enhancing the overall effectiveness of detection systems.

The standardized approach provided by our taxonomy fosters collaboration among researchers by creating a common terminology and framework for abusive language classification. This shared understanding can lead to more cohesive efforts, facilitate knowledge exchange, and accelerate progress in the field of online abuse detection and mitigation among researchers, policy makers, online platform owners, and other stakeholders.

Overall, our taxonomy serves as a valuable tool for addressing the multifaceted challenges of abusive language detection in online communications. By providing a standardized, flexible, and comprehensive framework for labeling and categorizing abusive language, it empowers stakeholders to improve model performance, manage the cost of dataset and model creation, and facilitate the communication between stakeholders.

## 8. Conclusion and Future Work

In this paper, we proposed a comprehensive taxonomy for abusive language, developed through an analysis of publicly available datasets published since 2020. Our hierarchical, faceted taxonomy includes dimensions such as context, target, intensity, directness, and theme, offering a flexible framework for labeling and classifying abusive language. The taxonomy not only provides a detailed classification system but also establishes a common terminology that enhances the reusability, interpretability, and comparability of datasets and models.

The evaluation of our taxonomy across eighteen datasets confirms its structural suitability in terms of generality, appropriateness, and orthogonality. By mapping 18 datasets to our taxonomy, we provide a mechanism for integrating the datasets, laying the foundation for improved dataset reusability and comparison. The primary practical applications of our taxonomy include facilitating dataset integration and comparison, enhancing model development, and fostering research collaboration.

As this represents the first comprehensive effort in abusive language taxonomy development, several areas for future work remain. Expanding the taxonomy to incorporate additional facets will further enhance its applicability. Conducting user studies regarding reliability and usability evaluations will provide insights into its applicability. Additionally, applying the taxonomy in empirical case studies will validate its effectiveness and support broader adoption.

**Acknowledgments**

We would like to express our sincere gratitude to Professor Frank Rudzicz for his invaluable feedback and insightful discussions. We also acknowledge the contributions of Si Cheng, for her assistance in preprocessing the datasets. The research was partially supported by an NSERC Discovery Grant.